\documentclass[letterpaper, 10 pt, conference]{ieeeconf}  

\IEEEoverridecommandlockouts                              

\usepackage[pages=some,placement=top, scale=1,
hshift=0,
vshift=-20,]{background}
\usepackage{lipsum}
\backgroundsetup{contents=This paper has been accepted for publication at IEEE CASE 2025,firstpage=true, angle=0,color=gray, opacity=1}

\usepackage{graphicx} 
\usepackage{subcaption} 
\usepackage[separate-uncertainty=true,multi-part-units=single]{siunitx} 
\usepackage{bm} 
\usepackage{amsmath} 
\usepackage{color}
\usepackage{colortbl}
\usepackage{soul} 
\usepackage{xcolor}
\usepackage[normalem]{ulem} 
\usepackage{float}
\usepackage{multirow}
\usepackage{comment}
\usepackage{amsmath}  
\usepackage{amssymb}  
\usepackage{placeins}
\usepackage{longtable, booktabs}
\UseRawInputEncoding
\usepackage{xcolor}
\usepackage{hyperref}

\newcommand{\rev}[1]{\textcolor{black}{#1}}

\title{\LARGE \bf
Multi-Class Human/Object Detection on Robot Manipulators using Proprioceptive Sensing
}

\author{Justin Hehli*$^{1}$, Marco Heiniger*$^{1}$, Maryam Rezayati$^{1\dagger}$, Hans Wernher van de Venn$^{2\dagger}$%
\thanks{$^{1}$J. Hehli, M. Heiniger, and M. Rezayati are with the Dep. of Informatics, University of Zurich, and MINDLab, Institute of Mechatronic Systems, ZHAW, Switzerland.
        {\tt\small justinsimon.hehli@uzh.ch}, {\tt\small marco.heiniger@uzh.ch}, {\tt\small rzma@zhaw.ch}}%
\thanks{$^{2}$H. W. van de Venn is with Institute of Mechatronics system, ZHAW, Switzerland
        {\tt\small vhns@zhaw.ch}}%
\thanks{Authors marked with *  contributed equally and with $\dagger$ are corresponding authors. This research was supported by the Eurostars project (Grant No. E!3087) titled \textit{SmartSenseAI}. We provide the code and dataset corresponding to this work at \href{https://github.com/MindLabZHAW/humanObjectDetection}{https://github.com/MindLabZHAW/humanObjectDetection} and \href{https://github.com/MindLabZHAW/humanObjectDetectionDataset}{https://github.com/MindLabZHAW/humanObjectDetectionDataset}}
}

\begin{document}

\maketitle
\thispagestyle{empty}
\pagestyle{empty}

\begin{abstract}

In physical human-robot collaboration (pHRC) settings, humans and robots collaborate directly in shared environments. Robots must analyze interactions with objects to ensure safety and facilitate meaningful workflows. One critical aspect is human/object detection, where the contacted object is identified. Past research introduced binary machine learning classifiers to distinguish between soft and hard objects. This study improves upon those results by evaluating three-class human/object detection models, offering more detailed contact analysis. A dataset was collected using the Franka Emika Panda robot manipulator, exploring preprocessing strategies for time-series analysis. Models including LSTM, GRU, and Transformers were trained on these datasets. The best-performing model achieved 91.11\% accuracy during real-time testing, demonstrating the feasibility of multi-class detection models. Additionally, a comparison of preprocessing strategies suggests a sliding window approach is optimal for this task.

\end{abstract}

\section{INTRODUCTION}\label{sec:introduction}

The field of human-robot collaboration (HRC) aims to provide shared workspaces where humans and robots work together collaboratively. This combination of human flexibility and cognitive ability with robot accuracy and endurance \cite{Mukherjee2022ASettings}  can improve flexibility and performance of production lines \cite{Popov2021Multi-ScenarioApplications}. Physical human–robot collaboration (pHRC) describes HRC settings where humans and robots come into direct physical contact with each other \cite{Ajoudani2018ProgressCollaboration, Park2023CollisionAlgorithms}. Ensuring human safety during these interactions is critical, especially where physical contact is an integral element of the workflow. To ensure safety, all physical interactions need to be identified and handled correctly by the robot through an appropriate contact/collision handling framework \cite{Park2023CollisionAlgorithms, Popov2023AdaptiveSensors}.

Supported by developments in the field of ML/AI and robotics, such contact interpretation pipelines include detection, recognition, and reaction stages \cite{Lippi2020EnablingReaction, Lippi2021ACollaboration}. Contact detection assesses whether any contact between the robot and an external object occurred \cite{Lippi2021ACollaboration, Fathi2022Human-RobotTasks}. The recognition phase consists of several objectives as follows:
\begin{itemize}
    \item Collision detection can be used to classify contacts into intentional contacts or unintentional contacts \cite{Al-Haija2022AsymmetricLearning}. 
    \item In case of an intentional contact, human intention detection may be done to ease communication between human and robot.
    \item In contact localization, the location of the contact on the robot body is identified \cite{Zwiener2018ContactLearning}. 
    \item Human/object detection identifies the contacted object, with the main goal of distinguishing humans and other types of objects.  
\end{itemize}
The reaction stage plans robot actions based on the output of the previous stages, for example interacting with a human or object, or stopping.

The focus of this paper is human/object detection. So far, to the best of our knowledge, the only relevant works investigating human/object detection are by Popov et al. \cite{Popov2021Multi-ScenarioApplications, Popov2023AdaptiveSensors, Popov2017CollisionSensors}. They use robot proprioceptive sensor data, such as joint torque or position, as model inputs. This constitutes a form of tactile perception, the process of interpreting touch sensing information to observe object properties \cite{Luo2017RoboticReview}. Tactile perception is essential for identifying an object's material properties, since vision alone cannot estimate such physical parameters on its own \cite{Luo2017RoboticReview}. The aforementioned studies classify the contacted objects into two classes, soft and hard, assuming soft contacts to be contacts with a human. This binary classification provides valuable insights, particularly regarding safety. However there are scenarios where a more detailed classification is beneficial or even necessary:

\begin{itemize}
    \item Many parts of the human body are rather hard and could be misclassified, with the robot falsely assuming the object isn't human. Training and predicting a dedicated human class may prevent this and improve safety.
    \item Soft (non-human) objects are likely to be assigned the soft class and assumed to be human, preventing the robot from properly interacting with such workpieces. Examples of this could be rubber or soft plastic objects.
    \item The single hard class essentially covers all non-human (hard) objects in the work cell, this limits the robot's decision making. Recognizing different objects could support a more sophisticated reaction stage, performing different actions based on the object type. For example, differentiating between metal and (hard) plastic objects could prove useful to decide whether the robot is touching its intended target or not.
\end{itemize}

Multi-class human/object detection models have the potential to solve many of these problems and support more complex applications. One such emerging application involves robots bodies manipulating objects, such as moving a component or guiding an object through a workspace, with a human operator potentially present in the work cell. In these cases, the ability to distinguish between a human, a soft object, and a hard object is critical for ensuring safety and appropriate responses. For instance, if a robot misclassifies a human as a non-human object, it may fail to take necessary precautions, compromising safety. \hypertarget{reviewer6_1}{\rev{While vision systems are often used, they may suffer from occlusions. Therefore, this system serves as a complementary sensation modality to enhance environmental awareness.}} On the other hand, distinguishing different materials enables the robot to fulfill its objectives, for example picking up a soft workpiece instead of a hard one. This context underscores the value of developing multi-class classification models that can reliably categorize contacts, and forms the basis of our research objective. However, prior work \cite{Popov2021Multi-ScenarioApplications} suggests that distinguishing between different soft objects is particularly challenging. To confirm or overcome this limitation, this paper investigates the feasibility of training such models. 

\textbf{Our contributions} are as follows:
\begin{itemize}
    \item \textbf{Development of a Three-Class Object Classifier:} We train a three-class object classifier that achieves a reasonable test accuracy (91.11\%). \hypertarget{reviewer5_1}{\rev{We decided to limit the number of classes to three in order to show a  proof-of-concept for multi-class classification in this context.}}In practice, the specific number and classes would depend  on the application context. For this study, we selected human, aluminum, and PVC. The human and aluminum classes correspond to the established soft and hard categories, while the PVC class serves as a middle ground in terms of stiffness and rigidity. We rely exclusively on proprioceptive sensor data to classify robot contacts.

    \item \textbf{Exploration of Data Collection and Preprocessing Strategies:} We examine various strategies and parameters for sensor data collection and preprocessing, analyzing their effects on model performance to identify optimal techniques. We highlight the best-performing strategies to support future research in robot-human/object contact recognition.

    \item \textbf{Implementation on a Real Robot:} We demonstrate the practical application of our multi-class classification approach by running it on a robot manipulator in real time. This confirms the model's real-world performance in dynamic environments.
    \item \textbf{Open Access to Code and Dataset:} The code and dataset used in this study are publicly available, facilitating reproducibility and encouraging further research in this area.
\end{itemize}


\section{Methods}\label{sec:methods} 

\subsection{Classification Models}
We trained and evaluated LSTM, GRU, and Transformer models. Recurrent LSTM and GRU models are well-established for sequence analysis tasks and thus an obvious choice. However, these recurrent models may struggle to capture long-range dependencies effectively \cite{Vaswani2017AttentionNeed, Hao2020ASeries}. For this reason, we also trained Transformer models, whose self-attention mechanisms may better capture long-range dependencies in our multivariate time-series data.

\subsubsection{LSTM and GRU Architecture}

\begin{figure}
    \centering
    \includegraphics[width=0.8\linewidth]{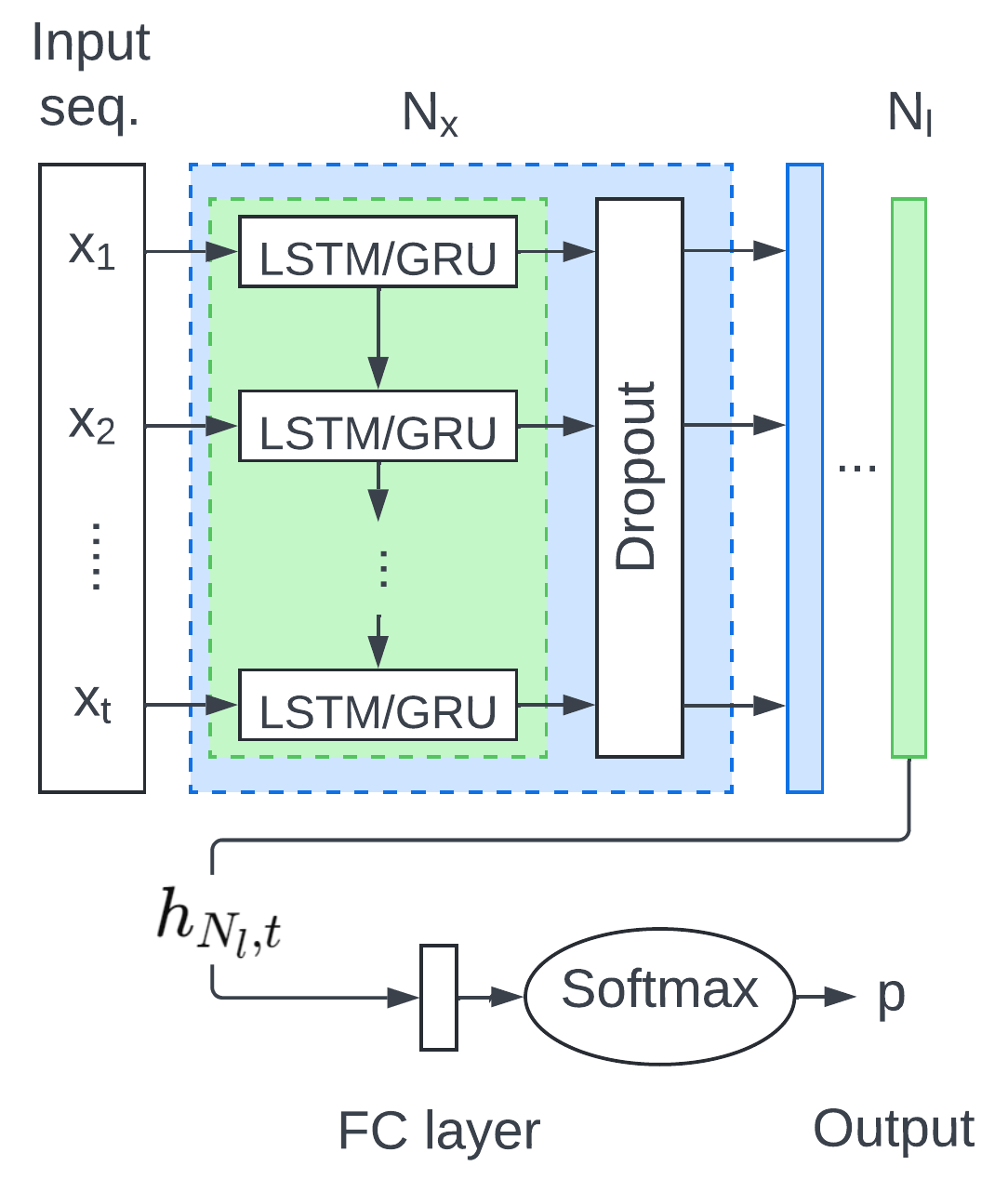}
    \caption{Recurrent network architectures: LSTM and GRU}
    \label{fig:rnn-architecture}
\end{figure}

As shown in Fig.\ref{fig:rnn-architecture}, the LSTM and GRU models consist of $N_l$ layers of $H_{cell}$ hidden size, with dropout layers of dropout probability $p_d$ after every recurrent layer except the last. This is followed by a fully connected layer with an input size of $H_{cell}$ and an output size of 3, which corresponds to the number of classes. The FC layer receives the hidden state of the last time step from the last recurrent layer as its input. Finally, class probabilities are calculated through Softmax.

\subsubsection{Transformer Architecture}
\label{transformer_arch_train}
\begin{figure*}
    \centering
    \includegraphics[width=0.8\textwidth]{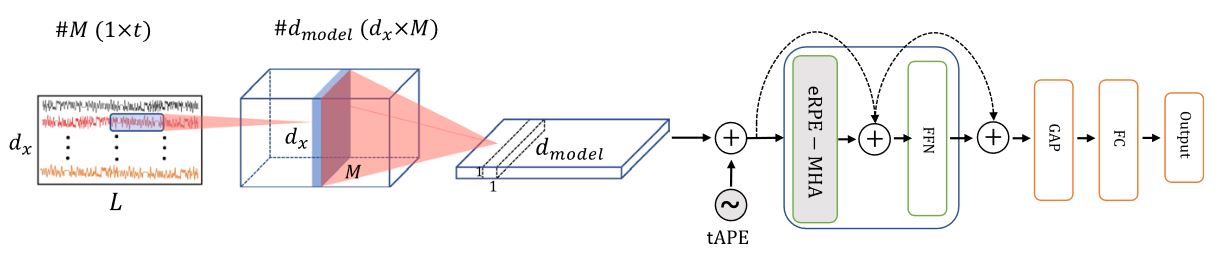}
    \caption{Time-series adapted encoder only Transformer architecture \cite{Foumani2024ImprovingClassification}. }
    \label{fig:transformer2}
\end{figure*}

The Transformer architecture \cite{Foumani2024ImprovingClassification} employed in this study, shown in Fig. \ref{fig:transformer2}, begins by processing the input time series data through convolutional layers to extract local temporal and spatial patterns. The first convolutional layer applies temporal filters to identify significant patterns within the series, while subsequent layers use spatial filters to capture correlations among variables in multivariate time series data. This reduces the sequence length and produces compact embeddings that are better suited for Transformer-based modeling.

To preserve global ordering information critical for time series analysis, it integrates Time Absolute Position Encoding (tAPE) into the model. As proposed by Foumani et al. \cite{Foumani2024ImprovingClassification}, tAPE incorporates the series length $L$, the embedding dimension $d_{model}$, and the positional index $k$. The encoding is designed to balance isotropy (making embeddings distinct) with distance-awareness (reflecting the relative positions of time points), ensuring that the positional relationships in time series are accurately captured. The tAPE is defined as:

\begin{equation} {\omega_k^{Foumani}} = \frac{ 10000^{-2k/d_{model}} \times d_{model}}{L} \label{eq:tAPE} 
\end{equation}

Additionally, to further enhance the model's ability to capture positional dependencies, it employs Efficient Relative Position Embedding (eRPE). Unlike absolute encodings, eRPE accounts for the relative shifts between positions $i$ and $j$ in the series. This enables the model to generalize better by emphasizing the relative relationships rather than fixed positional indices. The eRPE formulation is defined as \cite{Shaw2018Self-attentionRepresentations, Huang2020ImproveEmbeddings}:

\begin{equation}
\label{eq:eRPE}
    \alpha_i = \sum_{j \in L} \left( 
    \underbrace{\frac{\exp(e_{ij})}{\sum_{k \in L} \exp(e_{i,k})}}_{A_{i,j}} 
    + w_{i-j} \right) x_j
\end{equation}
Here, $L$ is the series length, $A_{ij}$ denotes the attention weight, $e_{ij}$ is the attention weight from position $j$ to $i$, $w_{i-j}$ is a learnable scalar representing the relative position weight, and $x_j$ is the input embedding.

Once the embeddings are enriched with tAPE and eRPE, they are fed into Transformer blocks where the multi-head attention mechanism processes long-range dependencies. These attention layers incorporate both the absolute and relative positional encodings to refine the representation of the series. The model concludes with pooling operations, including global average pooling and max-pooling, which summarize the learned representations into features suitable for classification.

This design, integrating convolutional layers with advanced positional encodings, enables the Transformer model to effectively capture both local and global patterns in multivariate time series data, resulting in improved classification accuracy \cite{Foumani2024ImprovingClassification}.

\subsection{Object Classification Dataset}
\label{subsection_datacollection}
\begin{figure}
    \centering
    \includegraphics[width=0.8\linewidth]{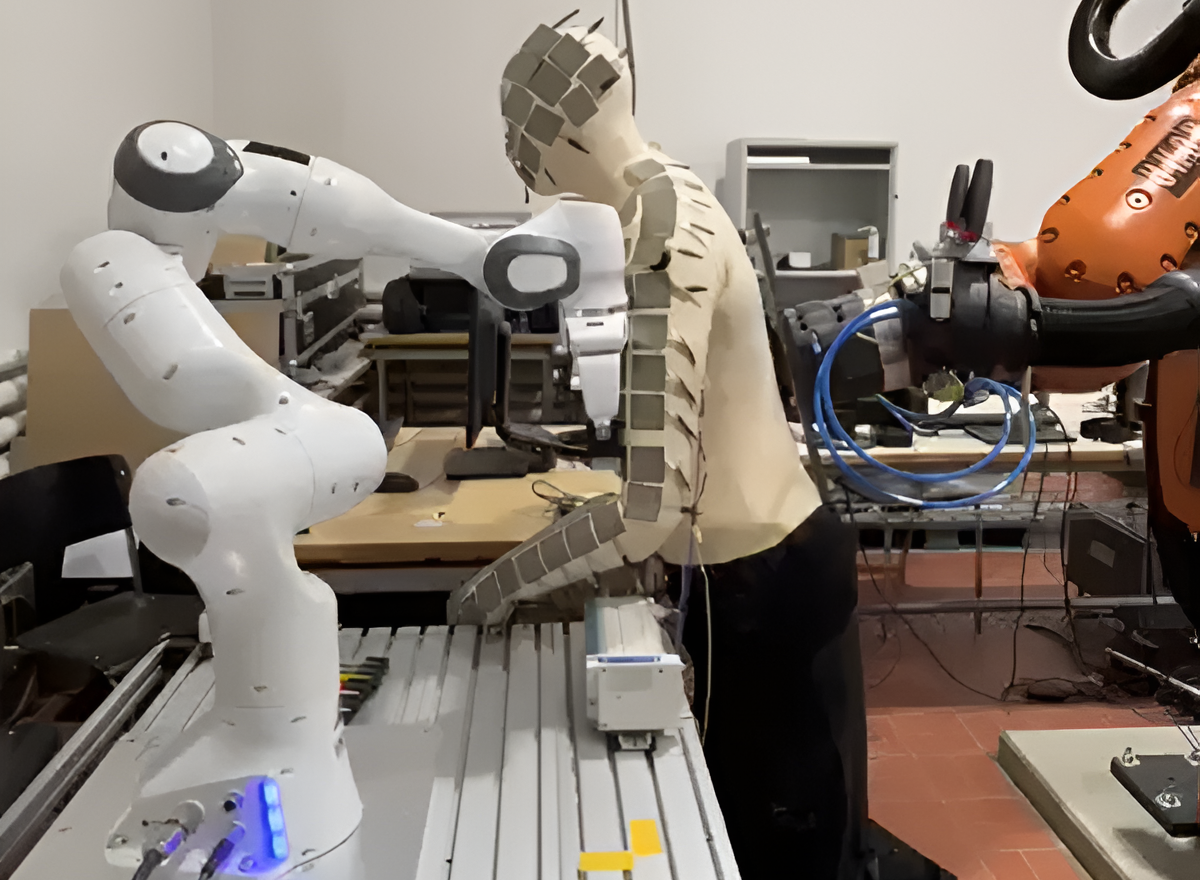}
    \caption{Data collection setup. The robot manipulator moves towards the dummy arm (or other objects) with the contact time recorded by sensors placed on the dummy body  (or other objects). 
    }
    \label{fig:data-collection-setup}
\end{figure}

In this study a dataset is designed and collected to train and validate models for the object classification task. The dataset consists of real-time data collected from a Franka Emika Panda robot manipulator interacting with three target objects. The dataset includes the following classes:

\begin{itemize}
   \item \textbf{Human Class}: Collisions with a dummy arm; \hypertarget{reviewer3_1}{\rev{Young's Modulus of the soft tissues $< 0.1$~GPa~\cite{Kalra2016Machanical}.}}
   \item \textbf{PVC Class}: Collisions with rigid PVC tube; \hypertarget{reviewer3_1}{\rev{Young's Modulus $\approx 3.4$~GPa~\cite{PolyvinylChlorideWikipedia}.}}
    \item \textbf{Aluminum Class}: Collisions with aluminum profiles; \hypertarget{reviewer3_1}{\rev{Young's Modulus $\approx 68$~GPa~\cite{6061AluminiumAlloyWikipedia}.}}

\end{itemize}
Data is recorded at a sampling rate of \SI{200}{\hertz}. The objects are strategically positioned to ensure a collision occurs during the robot's movement. A pressure-sensitive sensor, placed at the point of impact on each object, captures the contact times (Fig.\ref{fig:data-collection-setup}). To increase the variability in the  dataset, four to six different setup variations are implemented per motion, modifying either the object's location and/or the contact point on the robot link.

The final dataset comprises 85 time-series for training and 40 time-series for validation. Each time-series records multiple contacts between the robot and the respective objects, where 3 contacts per series were added to the dataset, as illustrated in Table \ref{tab:dataset_distribution}
\begin{table}[b]
    \centering
    \caption{Distribution of Contacts in the Dataset}
    \begin{tabular}{c c c}
        \hline
        \textbf{}       & \textbf{Training Set Contacts} & \textbf{Validation Set Contacts} \\
        \hline
        Human                 & 84                        & 54                            \\
        Aluminum              & 81                        & 27                            \\
        PVC                   & 90                        & 39                            \\
        \hline
        \textbf{Total}       & \textbf{255}              & \textbf{120}                  \\
        \hline
    \end{tabular}
    \label{tab:dataset_distribution}
\end{table}

\begin{figure}
    \centering
    \includegraphics[width=0.9\linewidth]{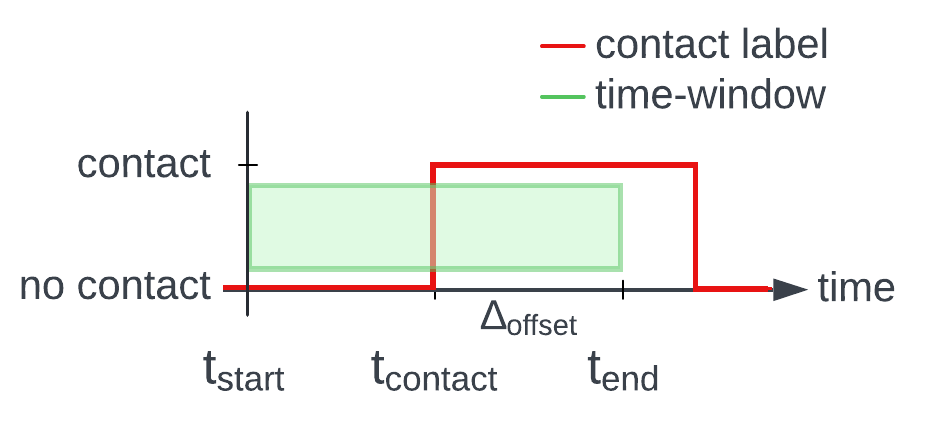}
    \caption{Illustrating the time-window. Data is processed by placing a window (200ms) 
    over the contact event ($t_{contact}$). $\Delta_{offset}$ adjusts the positioning of the extracted time-windows relative to $t_{contact}$.}
    \label{fig:contact-time-window}
\end{figure}

\subsection{Preprocessing and Model Input}
\label{subsection_preprocessing}
The human/object detection model is only inferred upon detecting a contact; therefore, only data around the contacts is relevant for training. Using a data windowing approach, patterns around each contact event are captured by extracting specific time-windows from the raw time-series data. As shown in Fig.\ref{fig:contact-time-window}, a time-window is defined from $t_{start}$ to $ t_{end}$. Each window consists of 40 data points, where each data point contains a vector of features as detailed in Section \ref{model_input}. With a sampling rate of $\SI{200}{\hertz}$, the window duration is $\SI{200}{\milli\second}$, 
where $t_{start} = t_{end} - \SI{200}{\milli\second}$.
We initially trained models exclusively on data collected during contacts, where $t_{start} = t_{contact}$. However, this approach resulted in low model performance during pilot testing. To improve performance, we introduced a preprocessing parameter $\Delta_{offset}$, which adjusts the positioning of the extracted time-windows relative to the contact time ($t_{contact})$ (Fig.\ref{fig:contact-time-window}), thereby modifying $t_{end}$ as follow:
\begin{equation}
    t_{end} = t_{contact} + \Delta_{offset}.
    \label{eq:t_end}
\end{equation}

\subsubsection{Fixed vs. Sliding Window Approach}
\begin{figure}
    \centering
    \includegraphics[width=0.8\linewidth]{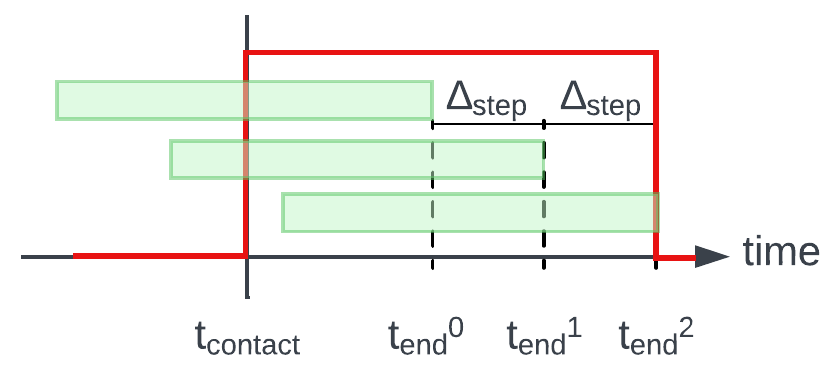}
    \caption{The sliding window approach with $\Delta_{step}$ step size.}
    \label{fig:sliding-contact-time-window}
\end{figure}

We further analyzed two different preprocessing approaches. In the \textit{fixed window} approach, a single time-window is extracted for each contact event. In contrast, the \textit{sliding window} approach segments data into multiple, overlapping windows that slide across the time-series, shown in Fig. \ref{fig:sliding-contact-time-window}. This method has been demonstrated in  \cite{Jaen-Vargas2022EffectsModels} to improve accuracy and other performance metrics. The window moves across the time-series with a step size of $\Delta_{step}$, starting at $t_{end}^0$ as follows:
\begin{equation}
    t_{end}^0 = t_{contact} + \Delta_{offset}
\end{equation}
\begin{equation}
    t_{end}^i = t_{end}^{i-1} + \Delta_{step} = t_{contact} + \Delta_{offset} + i \cdot \Delta_{step}
\end{equation}
with the condition  $t_{end}^i <= t_{contact} + \SI{300}{\milli\second}$, so only the first 300 ms of each contact are considered for capturing data. Thus, the number of extracted windows, and consequently the dataset size, depend on $\Delta_{offset}$ and the step size $\Delta_{step}$.

\begin{figure}
    \centering
    \includegraphics[width=0.8\linewidth]{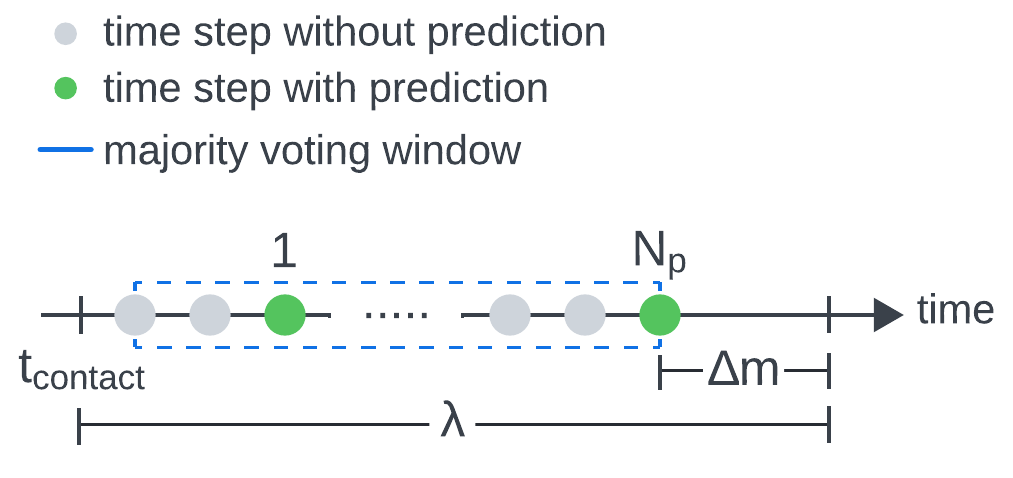}
    \caption{Majority voting with $N_p$ individual predictions, predicting at every 3rd time step,  resulting in a latency of $\lambda$.}
    \label{fig:majority-voting}
\end{figure}
\subsubsection{Model Input}\label{model_input}
The selected features are \textit{joint torque} ($\bm{\tau_J} \in \mathbb{R}^7$), \textit{joint position error} ($\bm{e_J} \in \mathbb{R}^7$), and \textit{joint velocity error} ($\bm{\dot{e}_J} \in \mathbb{R}^7$) of all 7 robot joints at every time step. The position and velocity errors are calculated as follows:

\begin{equation}
    \bm{e_{J,i}} = \bm{q_{J_{desired}, i}} - \bm{q_{J_{actual}, i}}
\end{equation}
\begin{equation}
    \bm{\dot{e}_{J,i}} = \bm{\dot{q}_{J_{desired},i}} - \bm{\dot{q}_{J_{desired}, i}}
\end{equation}
where $\bm{q} \in \mathbb{R}^7$ and $\bm{\dot{q}} \in \mathbb{R}^7$ are joint position and velocity, respectively. A model input element is given by one time-window as outlined before, and can be represented as
\begin{equation}
    X = \begin{bmatrix}
    \bm{e_J}^0 & \bm{\dot{e}_J}^0& \bm{\tau_J}^0\\
    \bm{e_J}^1& \bm{\dot{e}_J}^1& \bm{\tau_J}^1\\
    \vdots&\vdots& \vdots\\
    \bm{e_J}^{39}& \bm{\dot{e}_J}^{39}& \bm{\tau_J}^{39}
    \end{bmatrix}
\end{equation}
The used robot manipulator features seven joints, which results in 21 total input features. Model inputs are thus matrices of size $40\times21$.

In this work, we compare the performance of models trained on datasets containing either fixed windows or sliding windows of varying $\Delta_{step}$. Different $\Delta_{offset}$ 
values were also used in both fixed and sliding window approaches. We collected 85 time-series, each recording multiple contacts. Time-windows were extracted for three contacts per time-series, resulting in 255 analyzed contacts. Distinct datasets were generated for combinations of preprocessing hyperparameters, resulting in 16 training datasets of varying sizes, shown in table \ref{tab:dataset-sizes}. The number of data points per dataset may be reduced compared to its theoretical size due to data cleaning.
\begin{table}[b]
\centering
\caption{ Dataset sizes corresponding to preprocessing parameters}
\begin{tabular}{llll}
\hline
\multirow{2}{*}{$\mathbf{\Delta_{offset} (ms)}$} & \multicolumn{1}{c}{\multirow{2}{*}{\textbf{Fixed Window}}} & \multicolumn{2}{c}{\textbf{Sliding Window}}   \\
                                    & \multicolumn{1}{c}{}                                       & $\mathbf{\Delta_{step} = 1}$ & $\mathbf{\Delta_{step} = 4}$ \\ \hline
\textbf{5}                                   & -                                                          & 15093                 & 3774                  \\
\textbf{15}                                  & -                                                          & 14587                 & 3773                  \\
\textbf{25}                                  & 253                                                        & 14081                 & 3521                  \\
\textbf{50}                                  & 253                                                        & 12816                 & 3268                  \\
\textbf{75}                                  & 254                                                        & 11597                 & 3026                  \\
\textbf{100}                                 & 254                                                        & 10327                 & 2771                  \\ \hline
\end{tabular}
\label{tab:dataset-sizes}
\end{table}

Finally, we investigated multiple normalization methods, but found that all of them led to significantly worse performance, for all model types. Because of this, no normalization is being done during training and inference.


\begin{table*}[t]
    \centering
    \caption{Highest validation accuracy per model type, alongside the parameters used to achieve that accuracy}
    \begin{tabular}{llllll}
        \hline
        \textbf{Model} & \textbf{Accuracy} & $\mathbf{\Delta_{offset}}$ & $\mathbf{\Delta_{step}}$ & \textbf{Majority Voting} & \textbf{Dataset Size}\\ \hline
        GRU            & 88.70\%           & 50ms                  & 4                   & Hard, $N_p=15$         &3268  \\
        LSTM           & 92.17\%           & 5ms                   & 4                   & Hard, $N_p=8$          &3774  \\
        Transformer    & 93.04\%           & 15ms                  & 1                   & Hard, $N_p=15$         &14587  \\ \hline
    \end{tabular}    
    \label{tab:best-models-comparison}
\end{table*}

\begin{table}[t]
    \caption{Hyperparameters of top-performing models per model type (LR: Learning Rate, $p_d$: dropout probability) } 
    \centering
    \begin{tabular}{llllll}
        \hline
        \textbf{Model}  & \textbf{Epochs} & \textbf{LR} & $\mathbf{p_d}$ & \textbf{Model Specific} \\ \hline
        GRU               & 139         & 0.01        & 0.4            & \begin{tabular}[c]{@{}l@{}}$N_l=3$\\ $H_{cell}=40$\end{tabular}\\ 
        LSTM              & 107         & 0.01        & 0.4            & \begin{tabular}[c]{@{}l@{}}$N_l=2$\\ $H_{cell}=48$\end{tabular}                    \\ 
        Transformer       & 172         & 0.0001      & 0.2            & \begin{tabular}[c]{@{}l@{}}$d_{model}=8$\\ $\lambda_{L2}=0.2$\\ $h = 1$\end{tabular} \\ \hline
    \end{tabular}
    
    \label{tab:best-models-hyperparams}
    
\end{table}

\subsection{Majority Voting}
To enhance prediction stability, majority voting aggregates $N_p$ individual predictions.(Fig.\ref{fig:majority-voting}). 

Two different majority voting approaches, namely hard and soft voting, are used. In hard voting, each individual prediction contributes one vote, and the final prediction is the mode of all votes. In contrast, soft voting takes the average of the predicted probabilities and selects the class with the highest mean probability. Both methods are evaluated with different values of $N_p$ to assess their impact on classification accuracy.

To avoid blocking calls in the robotics software due to model runtime, models are only inferred at every third time step. With a time step period $T = \SI{5}{\milli\second}$, a range of $N_{p,min} = 8$ to $N_{p,max} = 15$, and an average model runtime of $\Delta_m = \SI{7.09}{\milli\second}$, the average prediction latency $\lambda$, ranging from $\lambda_{min}$ to $\lambda_{max}$, can be estimated as follows:
\begin{equation}
\lambda_{min} = N_{p,min} \cdot 3 \cdot T + \Delta_m \approx \SI{127.09}{\milli\second}
\end{equation}
\begin{equation}
\lambda_{max} = N_{p,max} \cdot 3 \cdot T + \Delta_m \approx \SI{232.09}{\milli\second}
\end{equation}

\subsection{Model Training and Hyperparameter Tuning} \label{Model Training}
For training LSTM and GRU models, we first split the (training) dataset into training and validation sets, at a ratio of 9:1. Then, k-fold cross validation with $k=5$ on the training set is used for tuning hyperparameters in a grid search approach. The final model is then trained with the optimal hyperparameter set found during cross validation. For both cross validation and final model training, the AdamW optimizer and cross-entropy loss are used. In the final model training, early stopping is implemented to combat overfitting. The trained models are then evaluated on the validation set, calculating prediction accuracy and confusion matrices.

For Transformer model training, we used the same dataset split as for the recurrent models. An exploratory hyperparameter search is conducted during parameter tuning. This process involves fine-tuning one parameter at a time while keeping the others fixed. This process is repeated for all parameters until all optimal model parameters, based on validation set performance, are found.  Similar to the LSTM and GRU models, early stopping and L2 regularization are applied to prevent overfitting.


\section{Experiments and Results}\label{sec:experiments_results} 

\subsection{Evaluation Methodology}\label{sec:evaluation-methodology}

The evaluation consisted of two distinct stages:

\begin{enumerate}
    \item \textit{Offline validation}:  This stage uses a separate validation set with preprocessing parameters configured to replicate real-time model inputs. Offline validation results also guide adjustments in preprocessing and refining the model hyperparameter range choices to improve training performance.
    \item \textit{Online testing} is conducted in real-time on the actual robot. Data is continuously collected, and upon detecting contact (via a secondary model or a contact sensor), the human/object detection model is inferred.
\end{enumerate}
\hypertarget{reviewer1_1,2}{\rev{For both offline validation and online testing, the same predefined motions as during training were used, but the placement of the contact object was altered randomly, in order to test on different, randomized data. The decision not to use new motions was due to time constraints, as the data collection process proved highly time-consuming and hard to set up for many movements. However, we believe this work still constitutes a valid proof-of-concept for the approach, even with limited test data.}}

Classification accuracy is the main performance metric throughout this study. 
\begin{equation}
    accuracy = \frac{correct\ predictions}{total\ predictions}
\end{equation}
Additionally, precision, recall, and a confusion matrix are employed to provide a comprehensive assessment.

\subsection{Offline Validation Results}
\label{sec:preprocessing-comparison}

\begin{figure*}
    \centering
    \begin{subfigure}{0.325\textwidth}
        \centering
        \includegraphics[width=\linewidth]{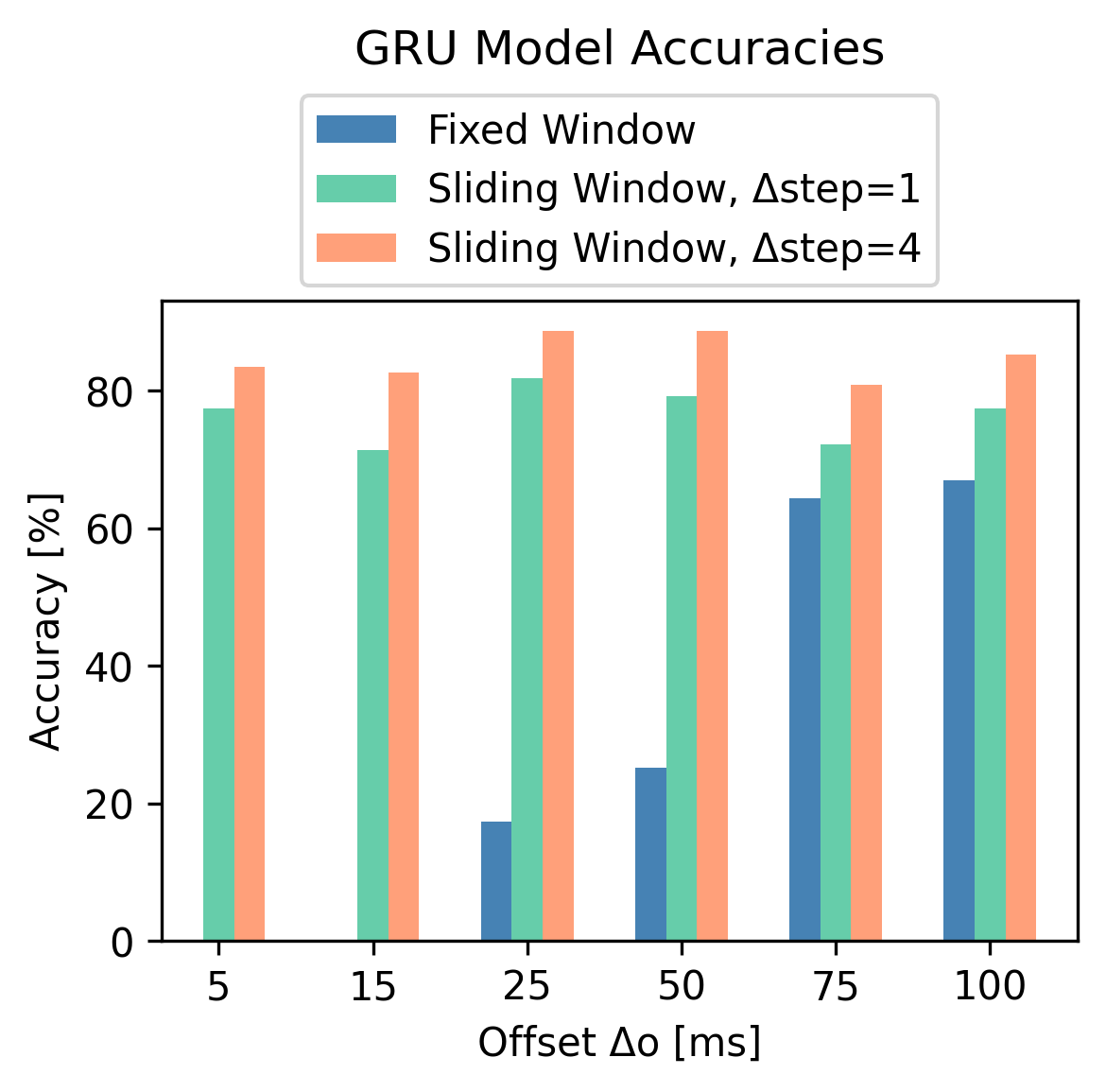}
        \label{fig:gru-accuracies}
    \end{subfigure}
    \begin{subfigure}{0.325\textwidth}
        \centering
        \includegraphics[width=\linewidth]{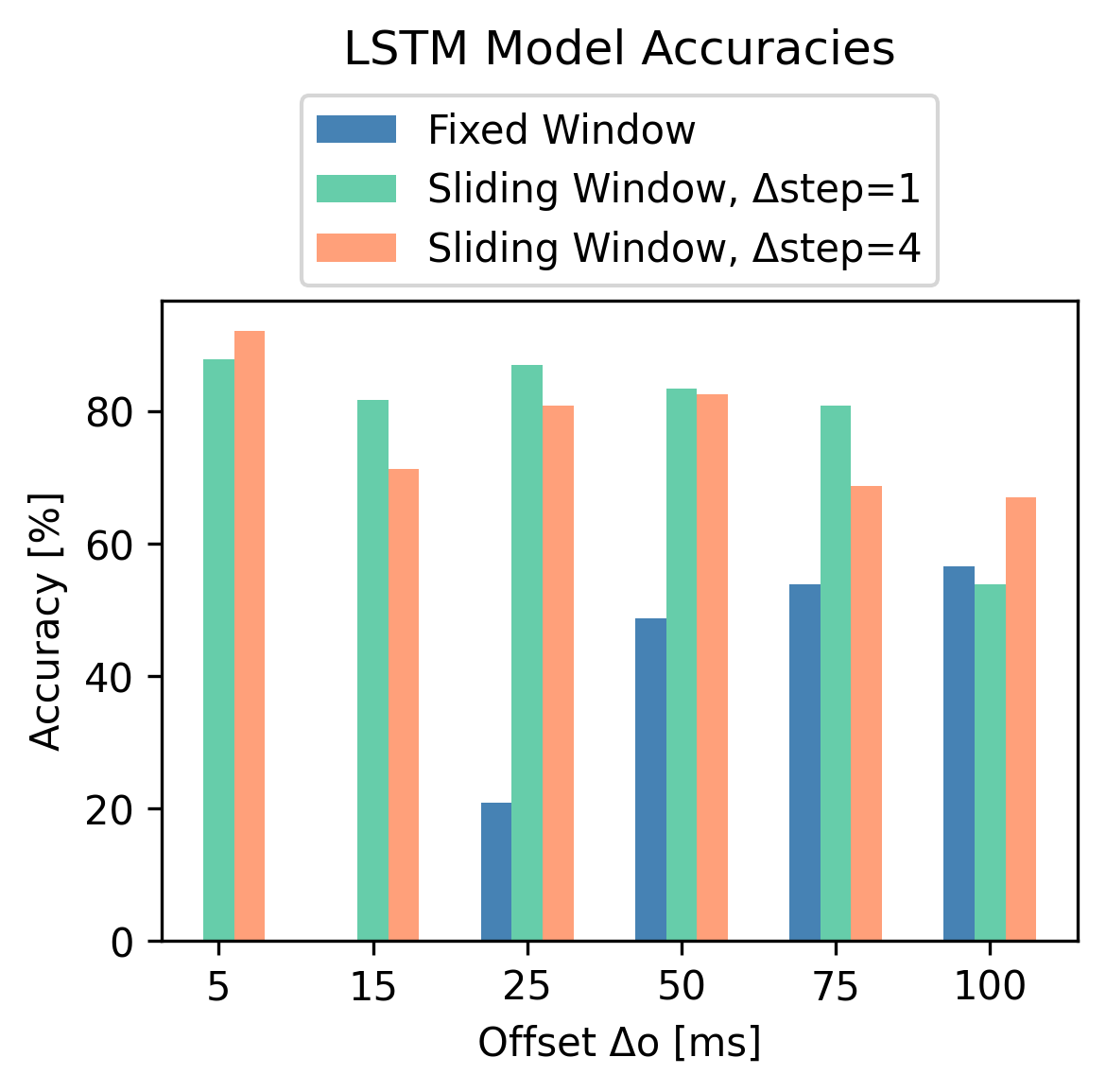}
        \label{fig:lstm-accuracies}
    \end{subfigure}
    \begin{subfigure}{0.325\textwidth}
        \centering
        \includegraphics[width=\linewidth]{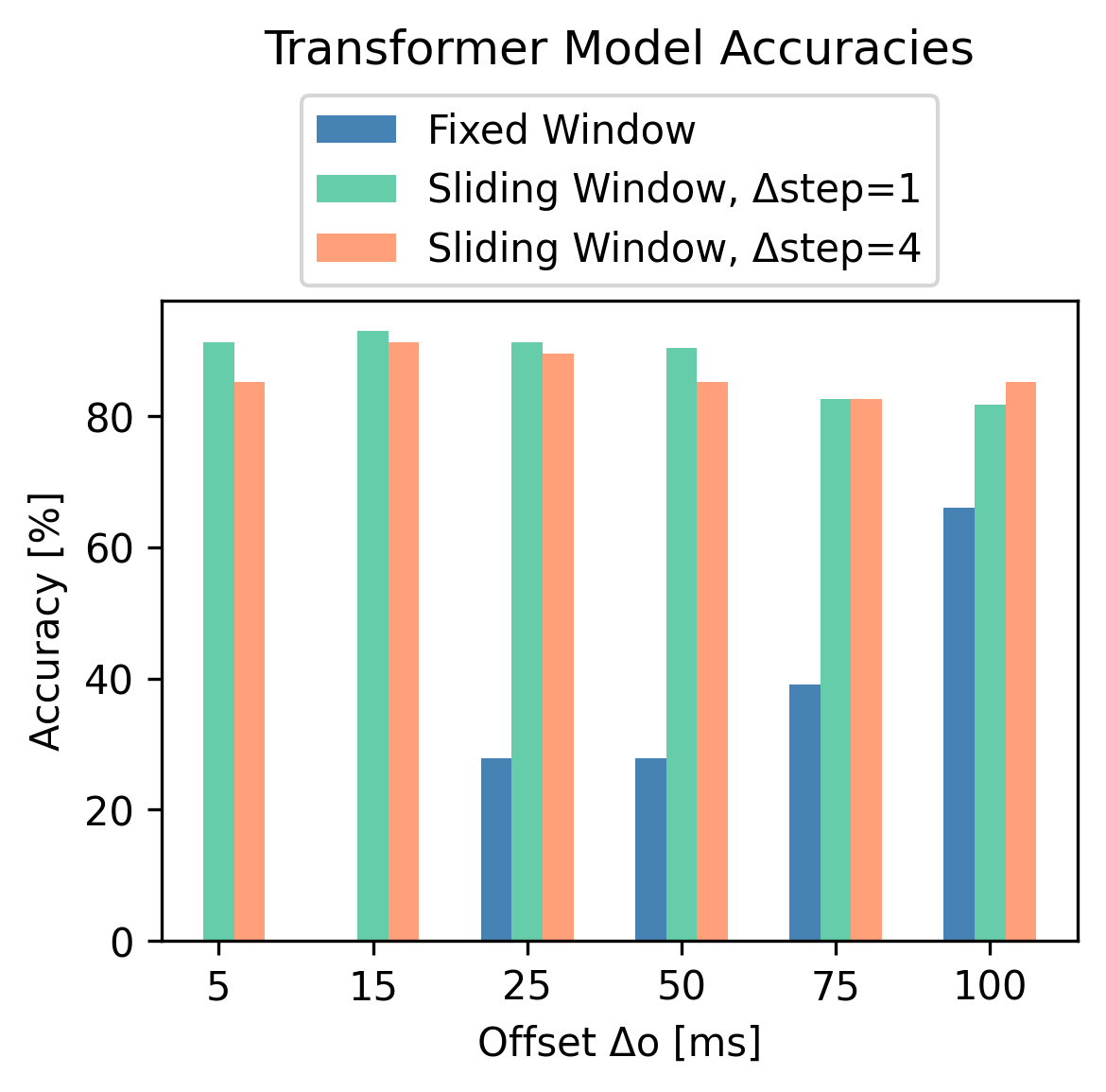}
        \label{fig:transformer-accuracies}
    \end{subfigure}
    \caption{Comparison of accuracies across preprocessing parameters and models}
    \label{fig:models-accuracies}
\end{figure*}

\begin{figure}
    \centering
    \includegraphics[width=\linewidth]{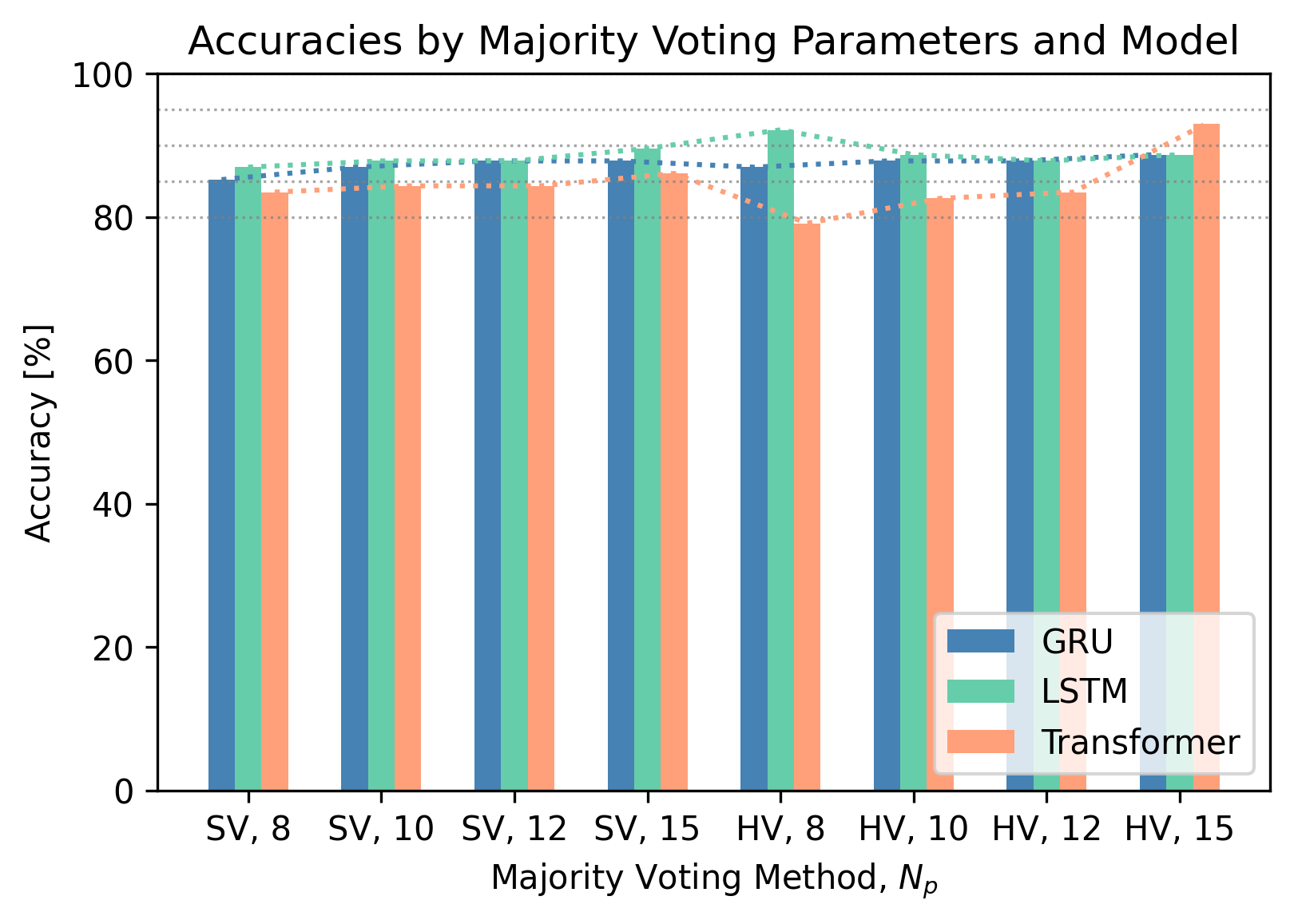}    
    \caption{\rev{Comparison of accuracies across majority voting parameters (SV: Soft Voting, HV: Hard Voting) and models. }}
    \label{fig:models-majority-voting}
\end{figure}

Based on the validation results, we compare model performance between different methods and parameters as follow:

\begin{itemize} 
    \item \textbf{Preprocessing methods and parameters}: we investigate the impact of using fixed vs sliding window preprocessing approaches, as well as different $\Delta_{offset}$  and $\Delta_{step}$ values on model performance.
    \item \textbf{Majority voting}: we analyze how implementing soft vs hard voting methods and the number of samples in majority voting affect model performance.
\end{itemize}

Table \ref{tab:best-models-comparison} presents the highest validation accuracy achieved by each model based on the parameters outlined above, while Table \ref{tab:best-models-hyperparams} details the hyper-parameters of these best models. Table \ref{tab:best-models-comparison} shows that training with datasets generated using the sliding window approach yields higher accuracy compared to the fixed window method across all model types, aligning with the findings of \cite{Jaen-Vargas2022EffectsModels}. The overall top-performing Transformer model achieved an accuracy of \SI{93.04}{\percent} on the validation set. Notably, hard majority voting outperforms soft majority voting across all three model types. However, the optimal number of predictions $N_p$ for achieving the highest accuracy differs among models.

Figure \ref{fig:models-accuracies}, compares the influence of pre-processing parameters on model accuracy. These parameters are fixed vs. sliding windows, $\Delta_{step}$ and $\Delta_{offset}$. We compare the accuracies of models trained on datasets with all possible pre-processing parameter combinations (see table \ref{tab:dataset-sizes}), in one plot per model type. To eliminate the influence of other parameters on this comparison, only data collected using the optimal majority voting strategy per model is used.

Figure \ref{fig:models-majority-voting}, then compares the influence of majority voting parameters, namely soft against hard voting and the number of considered individual predictions, $N_p$. For each model type, only validation data of the best model, as given in table \ref{tab:best-models-comparison} is used, in order to eliminate any influence of the pre-processing parameters on this comparison.

\subsection{Online Testing: Implementation on an Actual Robot}

\begin{table}
\centering
\caption{Precision, recall and f1-Score per class for online test results}
\begin{tabular}{llll}
\hline
                   & \textbf{Aluminum} & \textbf{PVC} & \textbf{Human} \\ \hline
\textbf{Precision} & 84.85\%            & 96.00\%      & 93.75\%        \\
\textbf{Recall}    & 93.33\%            & 80.00\%      & 100.00\%       \\
\textbf{F1-Score}  & 88.89\%            & 87.27\%      & 96.77\%        \\ \hline
\end{tabular}

\label{tab:transformer-precision-recall-f1}
\end{table}

\begin{figure}
    \centering
    \includegraphics[width=0.7\linewidth]{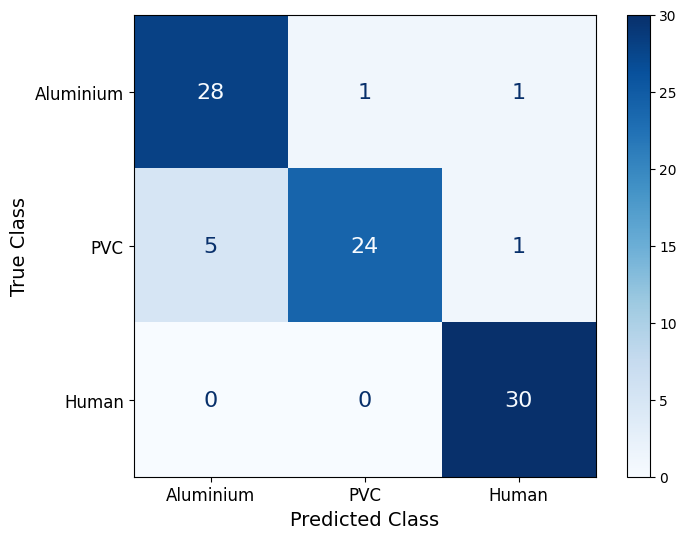}
    \caption{Confusion matrix for online test results}
    \label{fig:transformer-online-test-confusionmatrix}
\end{figure}

The best performing Transformer model, as detailed in Section \ref{subsection_preprocessing}, is tested in real-time on the robot manipulator to evaluate its performance in a practical setting. For each trained motion and class, 30 predictions were made and evaluated, resulting in an overall accuracy of 91.11\%. This accuracy is quite close to the results from the offline validation tests, indicating successful model deployment and implementation in a real-world environment. 

Class-wise precision, recall and f1-score are reported in Table \ref{tab:transformer-precision-recall-f1}, corresponding to the confusion matrix in Fig.\ref{fig:transformer-online-test-confusionmatrix}. The model's performance varies across different classes, as detailed below:
\begin{itemize}
    \item \textbf{Aluminum}: The precision of 84.85\% is the lowest among the three classes, indicating that the model has the highest difficulty identifying this class. Nevertheless, this is still a promising result, especially with a recall of 93.33\% for this class.
    \item \textbf{PVC}: The model demonstrates high precision at 96.00\%, reflecting its strong ability to correctly identify PVC instances, but it has a lower recall of 80.00\%, indicating that it fails to detect some PVC instances.
    \item \textbf{Human}: The model achieves perfect recall at 100.00\%, correctly identifying all Human instances, while its precision is also strong at 93.75\%.
\end{itemize}

These metrics highlight the strengths and weaknesses of the model in recognizing different materials and classes, providing valuable insights for potential improvements.


\section{Discussion}\label{sec:discussion}

\subsection{Impact of Preprocessing on Model Performance}

The analysis of preprocessing methods in \ref{sec:preprocessing-comparison} shows that models trained on sliding window datasets achieve significantly superior performance relative to those trained on fixed window datasets. This is in line with existing research \cite{Jaen-Vargas2022EffectsModels} and points towards the conclusion that sliding window preprocessing approaches are in general superior for time-series based human/object detection tasks. This performance difference could in part be due to the increased amount of data available when using sliding windows. Specifically in our case, with fixed windows approximately 255 samples are collected, whereas the sliding window method enables the construction of much larger datasets by utilizing overlapping segments.

For fixed window trained models, the accuracy visibly increases as \hypertarget{reviewer5_3}{\rev{$\Delta_{offset}$}} grows, seemingly tapering off above 75ms for GRU and LSTM models. This suggests that a larger offset allows models to focus on more relevant post-contact features, which improves their ability to make accurate predictions. However, no clear trend can be observed for models trained on sliding window datasets. While there is a trend for higher accuracy with $\Delta_{offset} \leq \SI{50}{\milli\second}$, in our opinion the results are not significant enough to make any final claims or suggestions regarding this parameter across models. The step size $\Delta_{step}$ does not seem to significantly influence model performance for the Transformer and LSTM models. In contrast, the GRU shows a clear tendency towards better performance with $\Delta_{step}=4$. This suggests that the GRU is able to learn overall data patterns effectively without requiring high redundancy from extensive overlap. Overall, while $\Delta_{step}$ may not significantly impact model performance across models, it does have a substantial effect on training time. Smaller $\Delta_{step}$ values increase the amount of training data, leading to longer training durations. Further research is necessary to either assert general recommendations regarding these preprocessing parameters in robot contact recognition tasks, or to prove that such parameters need to be tuned on a case-by-case basis. 

\subsection{Impact of Majority Voting Parameters}
Further notable parameters to consider are the majority voting method and its number of predictions $N_p$. In contrast to the pre-processing parameters discussed before, these parameters are only relevant during inference.

Overall, differences in performance between hard and soft voting approaches are quite low. However, all models performed slightly better with a hard voting approach. We can also observe that compiling more individual predictions, i.e. a higher $N_p$, usually leads to better accuracy, except in the case of LSTM, where hard voting with $N_p = 8$ performed best. Using some form of majority voting significantly stabilized the final performance of our predictions. The exact parameters should be evaluated per case, and will also be subject to requirements such as latency. 

\subsection{Model Performance and Feasibility}
The test accuracy of 91.11\% clearly demonstrates the feasibility of training multi-class human/object detection models. We can furthermore report a recall of 100.00\% for the human class. This is especially relevant because human operator safety is critical and therefore a high recall, or low false negative rate, is important for this class. 

Noticeably, the class-wise recall for PVC is slightly lower, with 80.00\%. All false negatives for this class occurred in two out of five motions during testing, which may hint towards some overfitting or hypersensitivity to certain motions or setups. This could potentially be due to a lack of variation in the data. We tested simpler models to address this issue, but their poor performance shifted our focus to more complex ones. Nevertheless, this work constitutes a viable proof of concept for three-class human/object detection models. In our opinion, future work should first focus on improving model robustness. To this end, a larger dataset with more motions, setup variations and contact locations would need to be collected and trained on to mitigate the dataset bias problem. This could hopefully increase the model's generalization capability, such that potentially, arbitrary motions and contact locations could be accurately analyzed, \hypertarget{reviewer5_1,2}{\rev{ which we see as the main limitation of our work. Another limitation is that we limited this work to three specific classes, as we aimed for a proof-of-concept of non-binary classification. Future work could also focus on training models with an increased number and/or different types of classes.}}


\section{Conclusion and Outlook}\label{sec:conclusion}

We collected a dataset for training human/object detection models and explored various preprocessing approaches and parameters. We then trained human/object detection models using Long Short-Term Memory, Gated Recurrent Unit and Transformer models and showed their ability to distinguish between three classes of contact objects. This serves as a proof of concept for multi-class human/object detection models. Based on our results, we suggest that future research should focus on improving model's generalization capabilities, mainly through collecting more training and test data. We furthermore showed that sliding window approaches are an effective preprocessing method that seems to outperform fixed window approaches. We could however not find any substantial trends regarding other preprocessing parameter's impacts on model performance, and suggest they be analyzed individually per case.

As stated, future work should first focus on improving model robustness \hypertarget{reviewer6_2}{\rev{and generalization performance}}. This could be accomplished by assembling larger training datasets on real-life robot manipulators. Alternatively, sim-to-real domain adaptation, where a model is pre-trained on simulation data and then fine-tuned on real-life data, could be investigated. The simulation data should be collected from an environment closely resembling the original real-world setup. This may be beneficial as it might reduce the need for real-life data collection, potentially leading to a huge decrease in time and effort. Lastly, our work only considers static objects that were fixed in place. Models that are able to distinguish static from dynamic (movable) objects would certainly prove useful in pHRC settings, especially regarding safety. Such models could potentially even be trained in conjunction with human/object detection models in a multitask learning setting.

\bibliographystyle{IEEEtran}
\bibliography{references}
\end{document}